\documentclass{article}
\usepackage{amsmath}
\usepackage{PRIMEarxiv}

\usepackage[utf8]{inputenc} 
\usepackage[T1]{fontenc}    
\usepackage{hyperref}       
\usepackage{url}            
\usepackage{booktabs}       
\usepackage{amsfonts}       
\usepackage{nicefrac}       
\usepackage{microtype}      
\usepackage{newunicodechar}
\newunicodechar{−}{\ensuremath{-}}
\usepackage{lipsum}
\usepackage{fancyhdr}       
\usepackage{graphicx}       
\graphicspath{{media/}}     

\hypersetup{
  colorlinks=true,
  linkcolor=black,
  citecolor=black,
  urlcolor=black
}

\title{Muscle Synergy Priors Enhance Biomechanical Fidelity in Predictive Musculoskeletal Locomotion Simulation}

\author{
  Ilseung Park\thanks{These authors contributed equally to this work} \\
  Department of Mechanical Engineering \\
  Carnegie Mellon University \\
  Pittsburgh, PA 15213, USA\\
  \texttt{ilseungp@andrew.cmu.edu} \\
   \And
  Eunsik Choi\footnotemark[1] \\
  Department of Physical Education \\
  Seoul National University\\
  Seoul, Republic of Korea\\
  \texttt{ces40320@snu.ac.kr} \\
     \And
  Jangwhan Ahn \\
  Lampe Joint Department of Biomedical Engineering\\
  UNC-Chapel Hill and NC State University\\
  Raleigh, NC 27695, USA\\
  \texttt{jahn26@ncsu.edu} \\
     \And
  Jooeun Ahn\thanks{Corresponding author} \\
  Department of Physical Education \\
  Seoul National University\\
  Seoul, Republic of Korea\\
  \texttt{ahnjooeun@snu.ac.kr} \\
}

\begin{document}
\maketitle

\begingroup
\renewcommand\thefootnote{}
\footnote{
Project page with supplementary videos: \url{https://ces40320.github.io/WebHomepage__Walk-RL}.
}
\endgroup

\begin{abstract}
Human locomotion emerges from high-dimensional neuromuscular control, making predictive musculoskeletal simulation challenging. 
We present a physiology-informed reinforcement-learning framework that constrains control using muscle synergies. 
We extracted a low-dimensional synergy basis from inverse musculoskeletal analyses of a small set of overground walking trials and used it as the action space for a muscle-driven three-dimensional model trained across variable speeds, slopes and uneven terrain.
The resulting controller generated stable gait from 0.7--1.8$~\mathrm{m\,s^{-1}}$ and on $\pm$ 6$^{\circ}$ grades and reproduced condition-dependent modulation of joint angles, joint moments and ground reaction forces.
Compared with an unconstrained controller, synergy-constrained control reduced non-physiological knee kinematics and kept knee moment profiles within the experimental envelope.
Across conditions, simulated vertical ground reaction forces correlated strongly with human measurements, and muscle-activation timing largely fell within inter-subject variability.
These results show that embedding neurophysiological structure into reinforcement learning can improve biomechanical fidelity and generalization in predictive human locomotion simulation with limited experimental data.
\end{abstract}

\keywords{Neuromuscular Simulation \and Muscle Synergy \and Deep Reinforcement Learning}

\section{Introduction}

Locomotion underpins human independence. The ability to walk, run, and maintain balance is closely linked to quality of life and autonomy, whereas declines in mobility are associated with increased risk of injury, hospitalization, and reduced well-being \cite{grimmer2019mobility, freiberger2020mobility}.
Understanding how locomotion is generated and sustained is therefore a central objective in both basic science and clinical research.
Achieving this objective requires a mechanistic account of locomotor control that goes beyond descriptive characterization toward causal explanations of how coordinated neuromuscular processes produce movement and how these mechanisms fail, adapt, or reorganize in response to pathology, training, or assistive interventions. 
Recent advances in musculoskeletal simulation provide a principled framework for such inquiry by integrating skeletal anatomy, muscle and tendon dynamics, and neural control into physics-based models. 
Nevertheless, the high complexity and redundancy of the human motor system pose significant challenges to developing controllers capable of reproducing human locomotion across diverse conditions.

Advances in reinforcement learning (RL) have enabled musculoskeletal simulations to generate diverse, human-like movements and to perform challenging locomotion tasks in complex, unstructured environments \cite{song2021deep}. 
Much of this progress, however, has been driven by motion imitation, where policies are trained to track pre-recorded motion-capture trajectories using motion-mimicking rewards \cite{simos2025reinforcement, park2025magnet, cotton2025kintwin}. 
While effective for reproducing the reference kinematics, imitation-based objectives primarily constrain surface-level motion and can leave the underlying neuromuscular coordination and joint kinetics underdetermined, allowing physiologically implausible internal solutions that weaken causal interpretability. 
Moreover, because these policies are optimized around the support of the demonstration distribution, they can be brittle under out-of-distribution conditions and exhibit limited generalization when speed, slope, or task demands depart from the motion clips.

To overcome these limitations, recent works have pursued predictive musculoskeletal simulation that generates human-like locomotion without explicit motion-imitation rewards from reference clips \cite{schumacher2025emergence, badie2025bioinspired, weng2021natural}. 
Broadly, two complementary strategies have emerged: shaping behavior with biologically plausible objectives \cite{schumacher2025emergence} and improving learning efficiency and robustness via bio-inspired curriculum design \cite{badie2025bioinspired, weng2021natural}. 
Despite this progress, it remains challenging to achieve quantitative agreement with human experiments, particularly beyond surface-level kinematics to joint kinetics and neuromuscular patterns across varying speeds and slopes. 
This persistent gap between visually plausible motion and experimentally grounded biomechanics motivates further work on controllers that are both physiologically constrained and broadly generalizable.

To address this gap, we adopt a biologically grounded prior based on muscle synergies.
One plausible contributor to the remaining discrepancy between model prediction and human behavior is the motor control strategy that has been overlooked. 
Machine learning in high-dimensional muscle control spaces admits many kinematically acceptable yet physiologically implausible internal solutions by allowing the independent control of individual muscles.
However, multiple evidences support that the central nervous system exploits low-dimensional control strategies, often conceptualized as muscle synergies, to coordinate large number of muscles during movement tasks. 
Experimental and computational studies in the field of motor neuroscience have shown that a small set of muscle synergies can effectively reproduce complex motor behaviors and simplify control without degrading performance \cite{berniker2009simplified, bizzi2013neural, bernstein1967coordination}. 
Electromyographic (EMG) analyses further confirm that muscle activation patterns can be decomposed into consistent low-dimensional modules across tasks \cite{davella2003combinations, clark2010merging, d2005shared, d2006control}.
Motivated by this observed principle in motor control, we extract a synergy basis from inverse musculoskeletal analyses of a small set of overground walking data and use it to constrain the action space of an RL policy. 
This process embeds neurophysiological structure into the controller, reduces the effective action dimensionality and discourages physiologically implausible solutions. 
Previous synergy-based action representations have indeed improved sample efficiency and robustness in over-actuated systems powered by muscle models \cite{berg2024sar, he2024dynsyn}. 
However, whether such representations translate into more human-like biomechanics across diverse conditions remains insufficiently established.

By integrating experimentally derived muscle synergies into RL-based predictive musculoskeletal simulation, we improve the quantitative fidelity of simulated locomotion across multiple speeds and slopes.
Specifically, the synergy-constrained controller produces gait that more closely matches experimental biomechanics---including joint kinematics and kinetics---than an unconstrained baseline while retaining robust performance under condition changes. 
These results demonstrate that incorporating neurophysiological structure into policy learning—by constraining control to a low-dimensional, physiology-informed action space—can narrow the gap between model predictions and human-like locomotion with more efficient control representations.

\section{Methods}

\subsection{Inverse Simulation: Muscle Synergy Extraction}

\subsubsection{Experiment}
One healthy young male (age: 28 years; mass: 69.56 kg; height: 176 cm) participated in this study. 
The participant had no prior history of neuromuscular or cardiovascular disorders. 
The study protocol adhered to the ethical principles of the Declaration of Helsinki and was approved by the Institutional Review Board (IRB) of Seoul National University (IRB No. 2408/003-008).

To collect kinematic and kinetic data, we utilized an optical motion capture system with twelve infrared cameras (OptiTrack Prime 13, NaturalPoint Inc., Oregon, USA) sampling at 100 Hz, and five force plates (four AMTI, Massachusetts, USA; one Bertec, Columbus, OH, USA) sampling at 1000 Hz.
A total of 38 reflective markers were affixed to the participant's anatomical landmarks.
The participant completed 10 trials of 10-meter overground walking, with one stride from each trial selected for subsequent analysis.
The collected marker trajectories and ground reaction force (GRF) data were processed using a fourth-order Butterworth low-pass filter with a cutoff frequency of 12 Hz.
Heel-strike events were defined as the time points at which the vertical GRF first exceeded 15 N.

\subsubsection{Muscle Activity Estimation}
We employed the musculoskeletal model named H2190, a 3D model with 21 degrees of freedom actuated by 90 muscles, including a lumbar joint and ten trunk muscles. This model, originally developed for the Hyfydy physics engine \cite{Geijtenbeek2021Hyfydy}, was converted for compatibility with OpenSim \cite{delp2007opensim}. The generic model was first scaled to the participant's anthropometry using the Scale Tool in OpenSim 4.5, based on marker trajectories from one walking trial. Subsequently, the inverse kinematics tool was used to compute joint angles throughout the walking motion.

To estimate the participant's lower-limb muscle activities, we utilized the MocoInverse tool within OpenSim Moco \cite{dembia2020opensim}. MocoInverse solves an optimal control problem that minimizes the sum of squared muscle control signals, alongside residual and reserve actuator efforts, while satisfying the constraints imposed by the measured joint kinematics and GRFs. This process yielded estimated activations for all muscles included in the model.

\subsubsection{Synergy Acquisition}
To create a low-dimensional controller, we performed dimensionality reduction on the estimated muscle activation matrix, $M \in \mathbb{R}^{T \times m}$, where $T$ is the number of time points and $m=40$ is the number of right-leg muscles.
Using non-negative matrix factorization (NMF) implemented via the scikit-learn toolbox, this matrix was decomposed into two non-negative matrices:
\begin{equation}
    M \approx WH,
\end{equation}
where $W \in \mathbb{R}^{T \times k}$ is the matrix of temporal activation coefficients, representing the time-varying activation patterns for each synergy, and $H \in \mathbb{R}^{k \times m}$ is the spatial synergy weight matrix, containing the constant weightings of each muscle within each synergy.
The dimensionality was reduced to ten synergies ($k=10$), resulting in a ${10\times40}$ synergy matrix $H$.
This matrix was then embedded as a low-dimensional controller in the predictive simulation framework.

\subsection{Predictive Simulation}

\subsubsection{Policy}
Two control representations were considered. 
For the synergy-constrained controller, the action space was defined as a linear combination of experimentally derived muscle synergies, whereas the independent controller directly controlled individual muscle excitations.
The agent's control policy was trained using the soft actor-critic (SAC) algorithm \cite{haarnoja2018soft}, provided by the Stable Baselines3 library \cite{stable-baselines3}. 
SAC is an off-policy, maximum entropy reinforcement learning framework designed to balance reward maximization with policy entropy, which encourages exploration and enhances robustness, particularly in complex, continuous action spaces.
The hyperparameters for the SAC agent were configured as follows.
The replay buffer size was set to 3,000,000, with learning commencing after an initial 10,000 steps of data collection. 
A minibatch size of 256 was used for training updates.
The learning rates for both the actor and critic networks were initialized at 0.001 and linearly decayed throughout training.
The agent was trained for a total of 75 million timesteps on an NVIDIA RTX 4090 GPU.
The policy network consisted of two hidden layers with 512, 512, and 256 units using ReLU activation, while the Q-function network comprised three hidden layers with 512, 512, and 256 units.
Training was conducted using five shared random seeds for both the independent and synergy-constrained controllers to ensure consistent initialization and fair comparison. 
Each seed was trained for 75 million environment interaction steps under identical hyperparameter settings. 
To characterize the maximal achievable performance of each control representation under equivalent training budgets, we selected the policy with the highest average episodic return measured during evaluation rollouts for subsequent biomechanical analysis.

\subsubsection{Simulation Environment}
The predictive simulations used the three-dimensional H2190 musculoskeletal model within the Hyfydy physics engine \cite{Geijtenbeek2019, Geijtenbeek2021Hyfydy}.
The agent received observations at 40 Hz, and each episode lasted 1000 steps, corresponding to 25 s.
To promote adaptation across walking speeds, training followed a deterministic curriculum of target velocities \([0.7, 0.8, \ldots, 1.6, 1.6, \ldots, 0.7]\) m/s applied sequentially across episodes rather than random sampling \cite{chiu2025learning}.
To expose the policy to diverse ground conditions, we generated 10{,}000 random terrains composed of sequential tiles with width 4 m, length 1 m, and height 0.1 m. The pitch of each tile was varied to produce slopes from \(-6^\circ\) to \(+6^\circ\).

At each control step, the agent generated a 30 dimensional action vector composed of 10 synergy activations for the right leg, 10 for the left leg, and 10 independent activations for the torso and pelvic muscles.
The synergy activations were multiplied by the synergy matrix \(H\) to produce full muscle activations for each limb.
To promote coordinated and symmetric motion, a phase-based mirroring strategy was applied such that the activation patterns of one side were mirrored to the opposite side \cite{abdolhosseini2019learning} when the model find the heel strike event. 

The observation vector comprised muscle fiber lengths and velocities, muscle forces, muscle excitations, head orientation represented as a quaternion, head angular velocity, the positions of both feet relative to the pelvis, joint angles and joint angular velocities, muscle activations, GRFs normalized by body weight, the center of mass velocity, and the current target forward speed.
To remove dependence on absolute position, the global base translational degrees of freedom were set to zero in the joint angle vector.

\subsubsection{Reward}
At each timestep the total reward is the weighted sum of four components:
\begin{equation}
    R \;=\; w_{\mathrm{vel}}\,R_{\mathrm{vel}} \;+\; w_{\mathrm{effort}}\,R_{\mathrm{effort}} \;+\; w_{\mathrm{rom}}\,R_{\mathrm{rom}} \;+\; w_{\mathrm{fall}}\,R_{\mathrm{fall}},
\end{equation}
where $w$ are scalar weights provided in the configuration and $R_{\mathrm{fall}}=1$ if the episode terminates due to a fall and $0$ otherwise.

The velocity component encourages stable forward progression while discouraging lateral drift and excessive head rotation. It is the product of three factors:
\begin{equation}
    R_{\mathrm{vel}} \;=\; R_{\mathrm{AP}} \, R_{\mathrm{ML}} \, R_{\mathrm{head}},
\end{equation}
with a plateaued Gaussian for tracking the forward center of mass speed $v_x$ around the current target $v_{\mathrm{target}}$,
\begin{equation}
R_{\mathrm{AP}} =
\begin{cases}
1, & \lvert v_x - v_{\mathrm{target}} \rvert \le \delta, \\[6pt]
\exp\!\Big(-c\,\dfrac{(\lvert v_x - v_{\mathrm{target}} \rvert - \delta)^2}{\sigma_{vx}^2}\Big), & \text{otherwise,}
\end{cases}
\end{equation}
a Gaussian penalty on the mediolateral velocity $v_z$,
\begin{equation}
    R_{\mathrm{ML}} \;=\; \exp\!\Big(-c\,\dfrac{v_z^2}{\sigma_{vz}^2}\Big),
\end{equation}
and an axis-wise Gaussian penalty on the head angular velocity $\boldsymbol{\omega}=(\omega_x,\omega_y,\omega_z)$,
\begin{equation}
    R_{\mathrm{head}} \;=\; \prod_{i\in\{x,y,z\}} \exp\!\Big(-c\,\dfrac{\omega_i^2}{\sigma_{\omega,i}^2}\Big).
\end{equation}
The gain is $c=0.06$, the forward-speed tolerance is $\delta=0.05$ m s$^{-1}$, the scale parameters are $\sigma_{vx}=0.07$ m s$^{-1}$ and $\sigma_{vz}=0.10$ m s$^{-1}$, and the angular-velocity scales are $(\sigma_{\omega,x},\sigma_{\omega,y},\sigma_{\omega,z})=(0.60,\,0.65,\,1.40)$ rad s$^{-1}$.

Control economy is encouraged through a quadratic effort term that sums muscle activations $a_i$ over all muscles,
\begin{equation}
    R_{\mathrm{effort}} \;=\; \sum_i a_i^2,
\end{equation}
which is assigned a negative weight so that larger activations decrease the total reward.

Biomechanical plausibility is enforced by penalizing range-of-motion violations at the knees and lumbar joint. 
Let $\theta_{\mathrm{knee},r}$ and $\theta_{\mathrm{knee},l}$ denote right and left knee angles, with hyperextension for $\theta>0$, and let $\theta_{\mathrm{lumbar}}$ denote lumbar extension. 
Using the upper-bound penalty $P_{\mathrm{up}}(x,u)=\max(0,\,x-u)$ and the box penalty $P_{\mathrm{box}}(x,\ell,u)=\max(0,\,x-u)+\max(0,\,\ell-x)$, the implemented term is set as
\begin{equation}
    R_{\mathrm{rom}} \;=\; P_{\mathrm{up}}(\theta_{\mathrm{knee},r},0) \;+\; P_{\mathrm{up}}(\theta_{\mathrm{knee},l},0) \;+\; P_{\mathrm{box}}(\theta_{\mathrm{lumbar}},-20^\circ,10^\circ),
\end{equation}
and is weighted negatively to discourage joint-limit violations. 
Subtalar limits were specified in development but are not active in the final computation.

Together these terms establish a trade-off among accurate speed tracking with stable head dynamics, energetic parsimony, adherence to range-of-motion limits, and robustness against falls, with the balance governed by the configurable weights.

\subsubsection{Evaluation: Comparison with Human Experimental Data}
We generated 10 evaluation rollouts for both  independent and synergistic controllers using the trained reinforcement learning policies under conditions identical to those used in two open-source experimental datasets \cite{scherpereel2023human, carmargo2021}, including level ground walking at the speed of 0.7 to 1.8 m/s and 5 degrees incline/decline walking at 1.2 m/s.
The final 10 strides from each rollout were used for comparative analysis.
To evaluate simulated locomotion across walking speeds, we used the treadmill dataset collected by Carmargo et al. from 22 young and healthy adults \cite{carmargo2021}.
All usable strides for each participant and condition were included following the procedures described in their original study \cite{carmargo2021}.
For the analysis of incline and decline conditions, we used the treadmill walking dataset collected by Scherpereel et al. \cite{scherpereel2023human}.
As hip rotation moments were unavailable for most participants, they were excluded from the analysis.
Eight participants without any additional data loss were included, and eight strides per participant were analyzed \cite{scherpereel2023human}.
Both simulation and experimental data were temporally normalized to 0–100\% of the gait cycle. 

To quantify agreement between simulated and experimental biomechanics, we evaluated joint kinematics (angles), joint kinetics (moments and GRFs), and muscle activities.
Root mean squared error (RMSE) ratio and Pearson correlation coefficients (r) were computed for each variable to serve as benchmarks.
Agreement was assessed using the RMSE ratio, which was defined as the RMSE between simulated and experimental patterns, normalized by the mean RMSE obtained from all cross-human comparisons within the experimental dataset. 
A ratio approaching 1.0 indicates that the error in the simulation is equivalent to the typical variability between individuals in the human dataset.
All evaluation metrics were computed separately for each locomotor condition for both independent and synergy-constrained controllers.

\section{Results}

\subsection{Reinforcement Learning Performance}

\begin{figure}[ht]
\noindent\resizebox{\textwidth}{!}{
\includegraphics[width=0.85\linewidth]{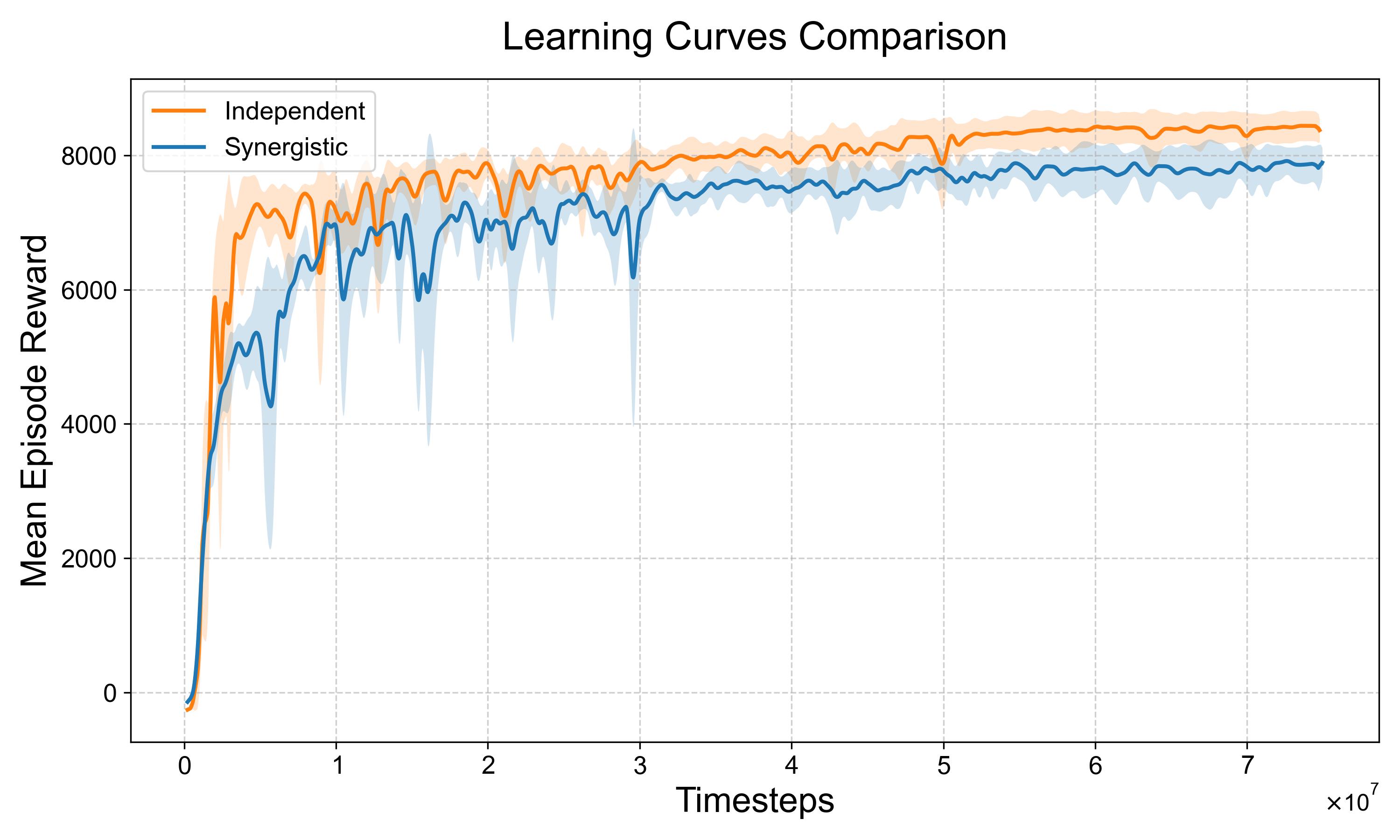}   
}
\caption{Comparison of reinforcement learning performance for independent and synergistic control strategies.
Mean episode reward is shown as a function of total training timesteps (×10$^7$), averaged over five training runs with different random seeds. Solid lines indicate a 10-point moving average applied to the mean rewards, and shaded regions represent the standard deviation across runs. Both control strategies exhibit stable learning behavior throughout training. The independent controller (orange) shows an earlier rise in reward and converges to a higher final performance level than the synergistic controller (blue).}
\label{fig:learning_curves}
\end{figure}

To assess learning performance, we analyzed training curves for two control strategies: independent and synergistic control.
Figure \ref{fig:learning_curves} presents the mean episode reward as a function of training timesteps, averaged over five runs with different random seeds. 
Shaded regions indicate the standard deviation across runs. A 10-point moving average was applied to all curves to reduce short-term fluctuations and highlight overall trends.

Both control strategies completed training without signs of instability or performance degradation under randomized uneven terrain and an identical speed curriculum.
Across repeated runs, the independent controller shows an earlier increase in reward and converges to a higher final performance level than the synergistic controller.
These trends are consistent across seeds, indicating reliable learning behavior for both control strategies.

\subsection{Similarity of Muscle Activities}

\begin{figure}[ht]
\noindent\resizebox{\textwidth}{!}{
\includegraphics[width=0.85\linewidth]{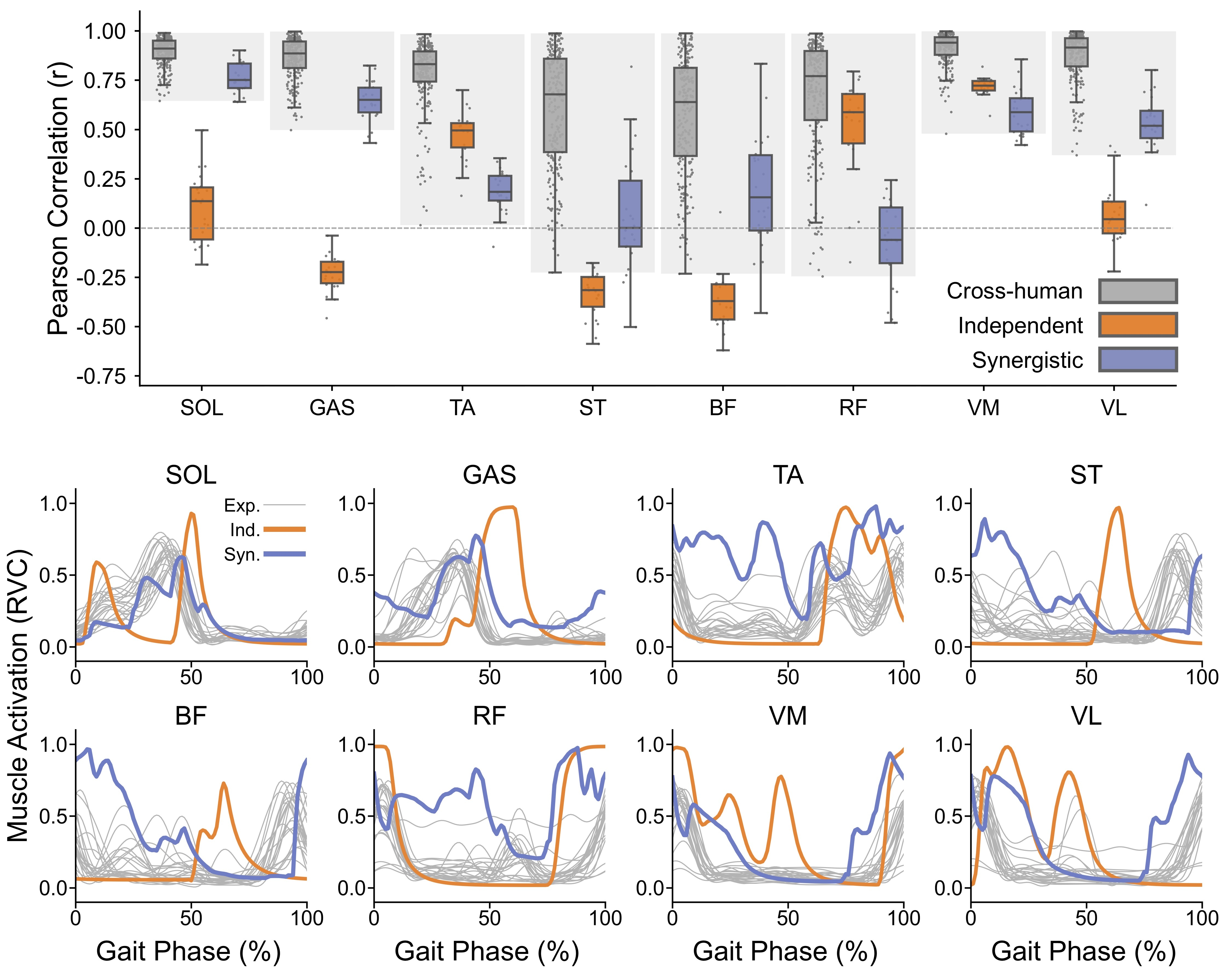}    
}
\caption{Similarity between predicted muscle activations and experimental EMG envelopes across eight lower-limb muscles.
Top box plots show Pearson correlation coefficients between predicted muscle activations and experimental EMG envelopes during level treadmill walking at 1.2 m/s for soleus (SOL), gastrocnemius medialis (GAS), tibialis anterior (TA), semitendinosus (ST), biceps femoris (BF), rectus femoris (RF), vastus medialis (VM), and vastus lateralis (VL).
Gray box plots represent the cross-human experimental benchmark derived from 22 participants of the open-source dataset, and the surrounding gray shaded regions indicate the corresponding envelope of cross-human variability.
Orange and blue box plots correspond to independent and synergistic control conditions, respectively.
Round dots denote individual correlation coefficients (n = 231 for cross-human pairwise comparisons; n = 22 per controller for correlations between simulated and experimental activations).
Bottom line plots show representative activation profiles over the normalized gait cycle, where gray curves indicate experimental EMG envelopes across participants, and colored curves represent predicted activations from the independent (orange) and synergistic (blue) controllers.
Muscle activations were normalized to reference voluntary contraction (RVC): for each muscle, predicted activations were scaled to the maximum amplitude observed within each controller, whereas experimental EMG envelopes were normalized to the subject-specific maximum amplitude.}
\label{fig:muscle_activity}
\end{figure}

The similarity between simulated muscle activations and experimental EMG envelopes was evaluated for eight major lower-limb muscles (soleus, SOL; gastrocnemius medialis, GAS; tibialis anterior, TA; semitendinosus, ST; biceps femoris, BF; rectus femoris, RF; vastus medialis, VM; vastus lateralis, VL).
Figure \ref{fig:muscle_activity} compares muscle activities from the independent and synergistic controllers against the cross-human variability benchmark under baseline walking conditions (1.2 m/s on a level treadmill). 
When evaluated against the same cross-human benchmark, the synergistic controller demonstrates consistently greater similarity to experimental EMG patterns than the independent controller.
Correlation values obtained using the independent controller stay outside the range of cross-human variability for five muscles (Figure \ref{fig:muscle_activity}), indicating limited overlap with experimentally observed activation profiles.
In contrast, the synergistic controller maintains activation patterns within the envelope of human variability for all muscles considered (Figure \ref{fig:muscle_activity}).
For the synergistic controller, ankle plantarflexors (SOL, GAS) exhibit strong agreement with human EMG patterns in both temporal structure and activation magnitude.
The ST, BF, and RF show lower correlations for the synergistic controller; these muscles also display broader cross-human variability relative to other muscles in the experimental dataset (Figure \ref{fig:muscle_activity}).

\subsection{Kinematic and Kinetic Similarity}

Within the same reinforcement learning framework, we trained separate policies for the independent and synergistic controllers, both of which successfully generate stable walking across all tested speeds (0.7–1.8 m/s) and slopes (−5°, 0°, +5°).
To examine the biomechanical plausibility of these controllers, we first examined whether each controller reproduces the directionality of speed- and slope-dependent modulation observed in the experimental data.

\begin{figure}[ht]
\noindent\resizebox{\textwidth}{!}{
\includegraphics[width=0.85\linewidth]{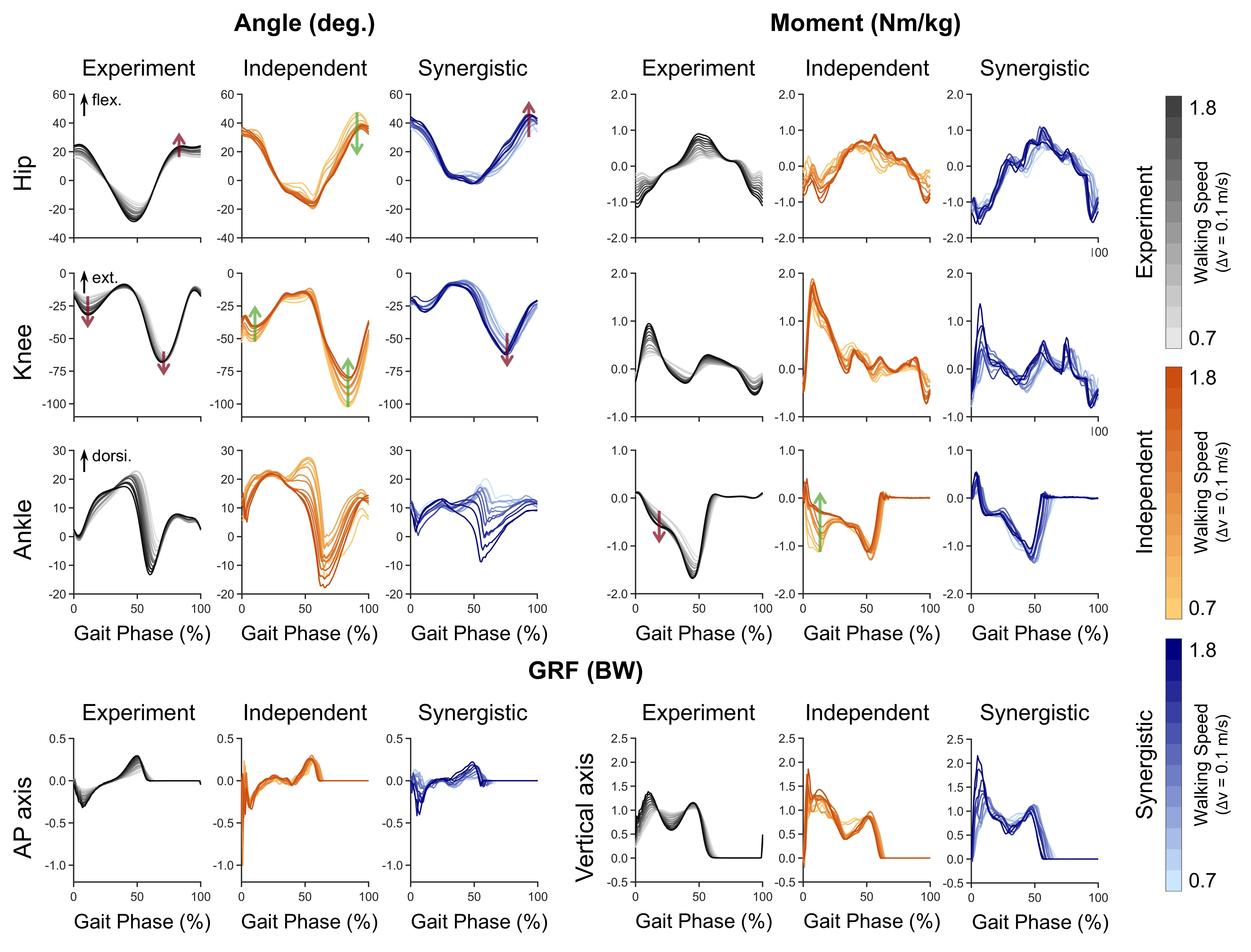}    
}
\caption{
Comparison of experimental and simulated kinematics, kinetics, and ground reaction forces across walking speeds.
Sagittal-plane joint angles and joint moments are shown for the hip, knee, and ankle over the gait cycle, alongside anterior–posterior (AP) and vertical GRFs.
Joint moments and GRFs were normalized by body mass.
Experimental measurements are shown in grayscale, independent control in orange, and synergistic control in blue.
For each condition, lighter colors indicate slower walking speeds (0.7 m/s) and darker colors indicate faster walking speeds (1.8 m/s) in increments of 0.1 m/s.
Arrows highlight gait phases exhibiting clear speed-dependent magnitude modulation.
Red arrows denote trends observed in the experimental data and similarly captured by the synergistic controller, whereas green arrows indicate regions where the independent controller deviates from the experimentally observed pattern.
}
\label{fig:speed_profiles}
\end{figure}

\begin{figure}[ht]
\noindent\resizebox{\textwidth}{!}{
\includegraphics[width=0.85\linewidth]{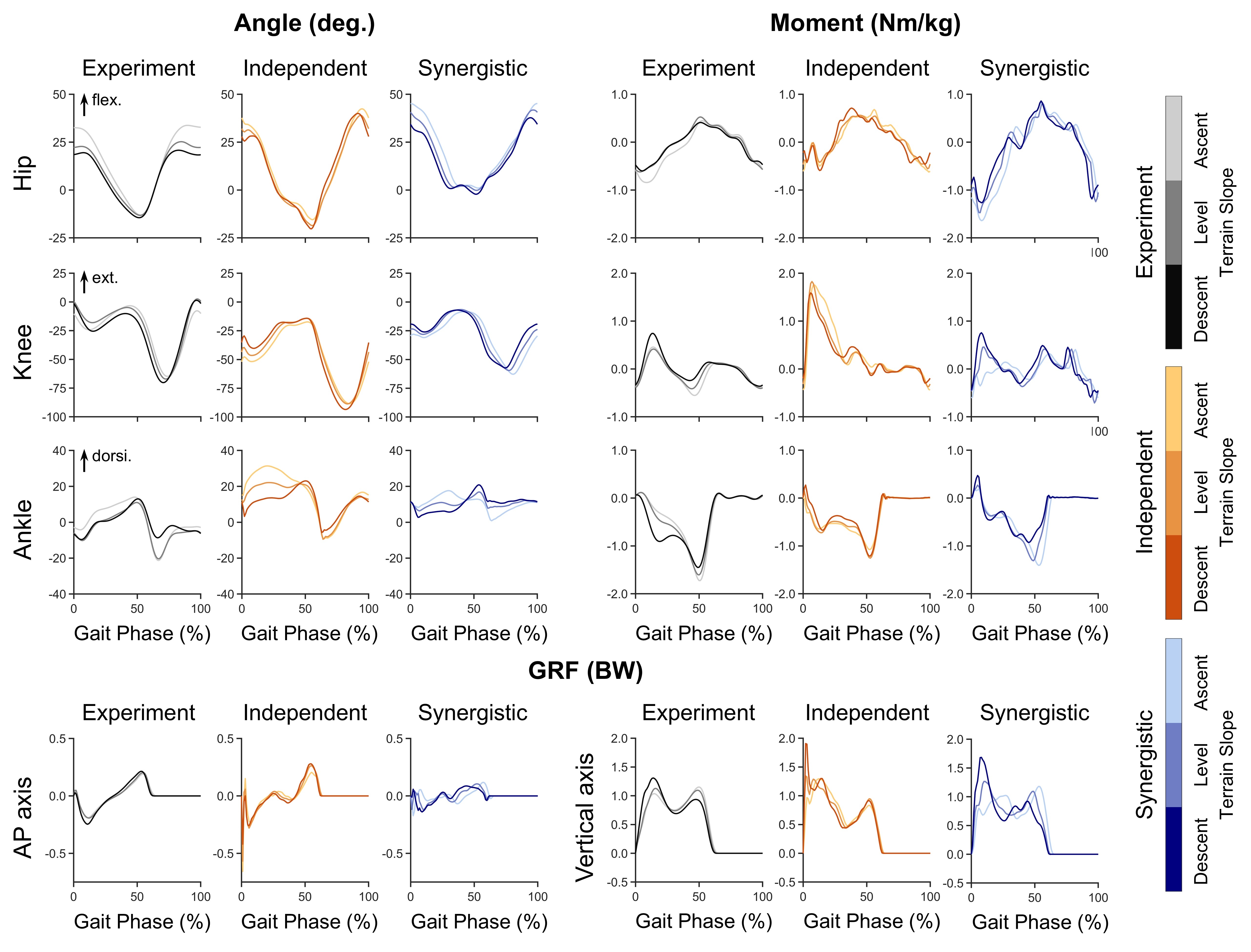}    
}
\caption{
Comparison of experimental and simulated kinematics, kinetics, and ground reaction forces across terrain slope conditions.
Sagittal-plane joint angles and joint moments are shown for the hip, knee, and ankle over the gait cycle, alongside AP and vertical GRFs.
Joint moments and GRFs were normalized by body mass.
Experimental measurements are shown in grayscale, independent control in orange, and synergistic control in blue.
For each condition, darker colors represent downhill walking (−5°), intermediate colors represent level walking (0°), and lighter colors represent uphill walking (+5°).
}
\label{fig:slope_profiles}
\end{figure}

Figures \ref{fig:speed_profiles} and \ref{fig:slope_profiles} show how closely the two simulated gait patterns reproduce the timing and magnitude of the experimentally observed lower-limb sagittal-plane joint angles, moments, and GRFs across walking speeds and slopes.
Across walking conditions, experimental joint angles and moments exhibit consistent monotonic magnitude changes across specific gait phases.
The physiology-informed synergy controller largely preserves these phase-specific trends. 
In contrast, the independent controller shows multiple instances of reversed or attenuated modulation, where the magnitude change across speeds occurs in the opposite direction relative to the pattern observed in experimental data. 
These trend inconsistencies are most evident in the hip angle during swing phase, knee angle during loading response and swing phase, and ankle moment during loading response phase (green arrows in Figure \ref{fig:speed_profiles}).

\begin{figure}[ht]
\noindent\resizebox{\textwidth}{!}{
\includegraphics[width=0.85\linewidth]{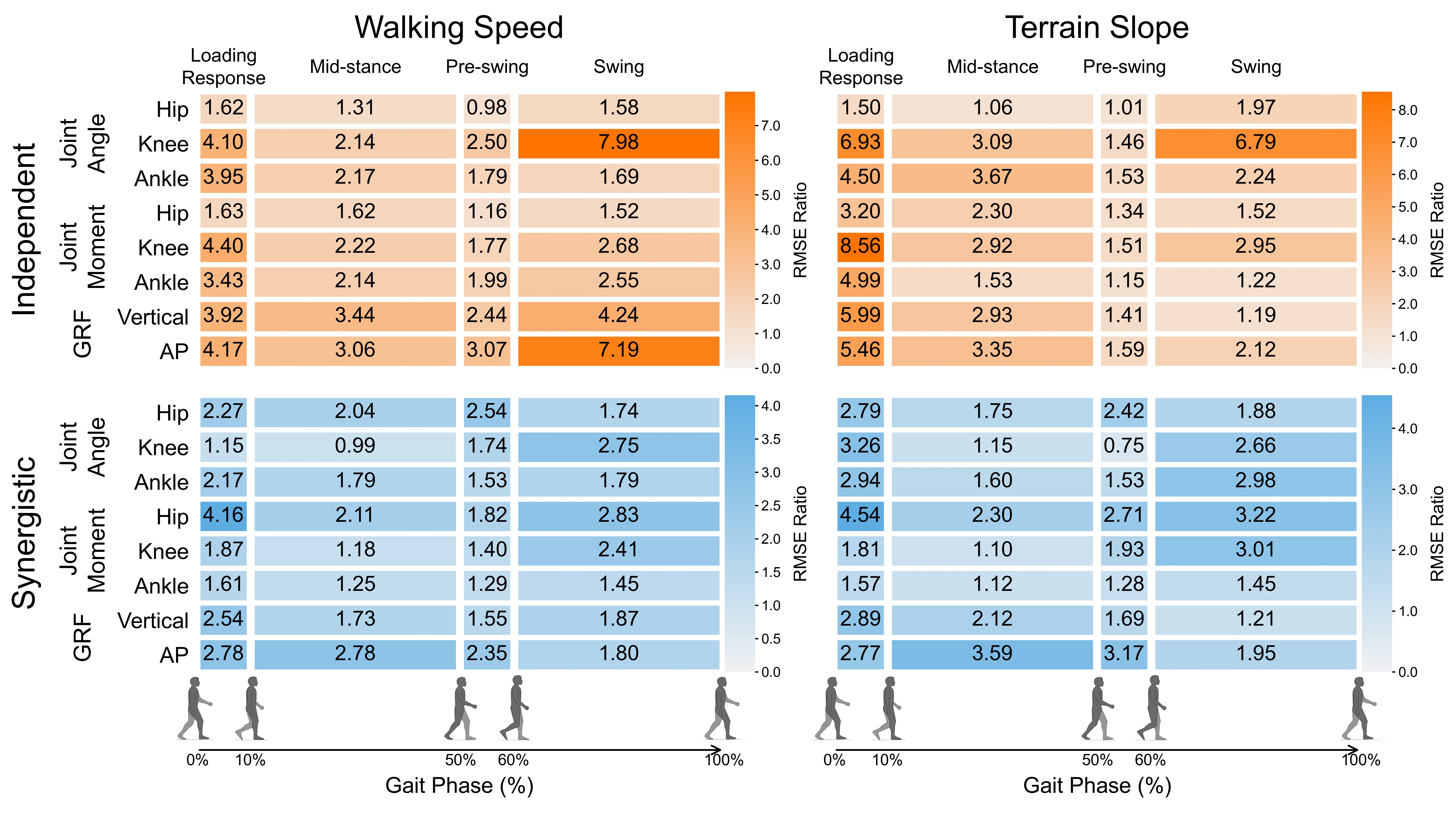}  
}
\caption{
Phase-dependent root-mean-square error (RMSE) ratios of simulated kinematics, kinetics, and GRFs relative to experimental data.
Heatmaps summarize the averaged RMSE ratios across gait phases (loading response, mid-stance, pre-swing, and swing) for independent control (orange) and synergistic control (blue) under both speed (left) and slope (right) conditions.
RMSE values for joint angles, joint moments, and GRFs were normalized by the corresponding cross-human RMSE; ratios close to 1.0 indicate the errors comparable to cross-human variability.
}
\label{fig:RMSE_heatmap}
\end{figure}

In addition to trend directionality, we evaluated overall waveform similarity using RMSE ratios.
Distinct patterns of agreement with the experimental data are observed between the independent and synergistic controllers, as summarized in Figure \ref{fig:RMSE_heatmap}.
Although both controllers exhibit instances of RMSE ratios exceeding 2.0 (39 for independent; 31 for synergistic), the divergence between controllers becomes pronounced at higher thresholds. 
Quantitatively, moderate-to-large deviations are substantially more frequent under independent control.
Across speed and slope conditions combined, 23 parameter–phase combinations exhibit RMSE ratios greater than 3.0 under independent control, compared with only 7 under synergistic control.
Extreme deviations (RMSE ratio > 5.0) occur exclusively under independent control (seven instances), whereas no such extreme values are observed under synergistic control (Figure \ref{fig:RMSE_heatmap}).

For the hip joint angle, the synergistic controller reproduces the detailed experimental trends but shows a larger global offset, resulting in moderately elevated RMSE ratios relative to the independent condition. 
However, for the knee and ankle joint angles, the independent controller exhibits substantially larger deviations. 
In particular, during the swing phase under speed conditions, the knee joint angle reaches an RMSE ratio of 7.98, representing the largest kinematic deviation across all joint angle variables. 
This extreme value coincides with a reversal of speed-dependent modulation in the independent controller.

Joint moment patterns show a similar discrepancy between the two controllers. 
For the synergistic controller, the largest kinetic deviation occurs in the hip flexion moment during loading response phase (RMSE ratio of 4.16 for walking speed; 4.54 for walking slope). 
In contrast, the independent controller produces substantially larger errors at the knee joint during the same phase. 
Under slope conditions, the knee moment resulting from the independent controller during loading response phase reaches an RMSE ratio of 8.56, which is close to double the maximum deviation observed in joint moments in the synergistic controller. 
For ankle moments, the synergistic controller shows condition-specific variations at lower speeds but remains bounded, with a maximum RMSE ratio of 1.61, whereas the independent controller consistently exceeds this level.

Both controllers reproduce the fundamental GRF structure, including the braking-to-propulsion pattern in the anterior–posterior (AP) axis and the characteristic double-peak profile in the vertical axis. 
However, larger deviations are observed under the independent control. 
In particular, delayed toe-off timing in the AP axis of GRF under speed conditions contributes to an RMSE ratio of 7.19 during swing phase, representing one of the largest force-related deviations. 
Although loading response phases exhibit elevated RMSE ratios under both controllers, the magnitude and frequency of extreme deviations are consistently greater under the independent control.

\section{Discussion}

This study shows that embedding experimentally derived muscle synergies into a reinforcement-learning controller improves the quantitative fidelity of predictive musculoskeletal simulations across walking speeds and ground slopes, without relying on motion-trajectory imitation during training.
By constraining actions to a low-dimensional synergy space extracted from inverse musculoskeletal analyses and non-negative matrix factorization, the policy incorporates neurophysiological structure into control and yields simulated kinematics, kinetics, and ground reaction forces that more closely match human data compared to the conventional independently controlled musculoskeletal model.

Relative to prior predictive musculoskeletal-RL studies, our work advances both model fidelity and the breadth of experimental validation.
Weng et al.~\cite{weng2021natural} demonstrated close-to-natural walking in a constrained sagittal-plane setup with a relatively low-dimensional musculoskeletal model, reporting kinematic agreement (joint-angle correlations) and spatiotemporal metrics but they provided limited kinetic validation beyond motion-level similarity.
Badie et al.~\cite{badie2025bioinspired} moved to a 3D model with a compact muscle set per leg and explicitly compared simulated joint kinematics and GRFs to human data, highlighting remaining discrepancies (e.g., ankle and knee features and GRF shape), but they did not explicitly benchmark joint moment profiles or EMG agreement.
Schumacher et al.~\cite{schumacher2025emergence} further demonstrated robustness across multiple musculoskeletal models spanning low- to high-dimensional actuation, and evaluated similarity using experimental-match metrics based primarily on lower-limb joint angles and GRFs, but still without direct EMG envelope comparisons.
In contrast, using a full 3D, high-muscle-count model, we quantitatively benchmark not only kinematics and GRFs but also joint kinetics and muscle-activity patterns against human experimental references across a wide range of speeds and slopes, showing that synergy-constrained control improves biomechanical fidelity at both the joint and muscle levels.

The value of synergy-constrained control may be even greater when simulating pathological locomotion, particularly in neurological disorders where gait impairment reflects altered neuromuscular coordination rather than kinematics alone.
Because similar joint trajectories can be produced by many different muscle-level coordination strategies, approaches that primarily match motion may reproduce the appearance of impairment while leaving the underlying pathological control mechanisms unrevealed.
In post-stroke walking, for example, EMG-based synergy analyses consistently report fewer and less distinct modules in the paretic limb, with evidence of increased co-activation and altered activation timing \cite{clark2010merging, van2020lower, defour2025impaired}.
Critically, the number and structure of these modules are associated with lower-extremity motor function and walking performance \cite{clark2010merging, van2020lower, defour2025impaired}.
These findings motivate a natural extension of our framework: by replacing the healthy synergy basis with patient-specific (or impairment-graded) synergy structure—potentially including reduced dimensionality—predictive simulation could better reflect the neuromotor constraints that give rise to pathological gait.
Such physiology-informed simulations could support mechanistic hypothesis testing and provide a principled substrate for designing and evaluating assistive-device controllers and rehabilitation strategies in clinical settings.

Several limitations should be acknowledged.
First, we evaluated only walking on level ground and modest slopes, so running, turning, and stair ambulation remain untested.
Second, the synergy basis was obtained from inverse musculoskeletal simulation rather than directly from electromyography.
Although recording EMG from all relevant muscles is not feasible, future work could explore EMG-driven models or hybrid approaches that fuse partial EMG with inverse estimates.
Third, the synergy set and its dimensionality were fixed across tasks and speeds, which may restrict adaptation; allowing state dependent or subject-specific synergies may improve fidelity.
Fourth, the approach was calibrated with data from a single participant, which may limits generalization across populations with different anthropometrics and motor strategies.
Nonetheless, the framework improved biomechanical fidelity across speeds and slopes even when the synergy basis was derived from a single individual, indicating robustness.
Extending the framework to broader locomotor tasks, physiologically grounded or adaptive synergy formulations, and larger, more diverse cohorts will be necessary to fully validate the generalizability and predictive scope of the devised framework.

\section{Conclusion}
Our findings show that physiology-informed action representations can produce human-like locomotion in predictive simulation with minimal experimental data.
Constraining control to synergies derived from inverse analysis enables encapsulation of human kinematics, kinetics, and GRFs across speeds and slopes while using only a high level velocity target.
This establishes a practical path toward sample efficient and mechanistically grounded controllers for rehabilitation, assistive devices, and exoskeleton control.
Future work will extend beyond level walking, incorporate subject diverse datasets and EMG driven or hybrid synergy estimation, allow state dependent or adaptive synergy bases, and include uncertainty aware evaluation and prospective validation on hardware. 
We expect to advance both causal understanding of locomotion and the design of assistive technologies by leveraging not only human-like musculoskeletal structure but also human-like control strategy, which has been discovered in motor neuroscience.

\bibliographystyle{unsrt}  
\bibliography{references}

@article{ Geijtenbeek2019,
  author = {Thomas Geijtenbeek},
  title = {SCONE: Open Source Software for Predictive Simulation of Biological Motion},
  journal = {Journal of Open Source Software},
  year = {2019},
  volume = {4},
  number = {38},
  pages = {1421},
  publisher = {The Open Journal},
  doi = {10.21105/joss.01421},
  url = {https://doi.org/10.21105/joss.01421},
}

@misc{Geijtenbeek2021Hyfydy,
  author = {Geijtenbeek, Thomas},
  title = {The {Hyfydy} Simulation Software},
  year = {2021},
  month = {11},
  url = {https://hyfydy.com},
  note = {\url{https://hyfydy.com}}
}

@inproceedings{chiu2025learning,
  title={Learning speed-adaptive walking agent using imitation learning with physics-informed simulation},
  author={Chiu, Yi-Hung and Lee, Ung Hee and Song, Changseob and Hu, Manaen and Kang, Inseung},
  booktitle={2025 International Conference On Rehabilitation Robotics (ICORR)},
  pages={835--841},
  year={2025},
  organization={IEEE}
}

@inproceedings{abdolhosseini2019learning,
  title={On learning symmetric locomotion},
  author={Abdolhosseini, Farzad and Ling, Hung Yu and Xie, Zhaoming and Peng, Xue Bin and Van de Panne, Michiel},
  booktitle={Proceedings of the 12th ACM SIGGRAPH Conference on Motion, Interaction and Games},
  pages={1--10},
  year={2019}
}

@article{grimmer2019mobility,
  title={Mobility related physical and functional losses due to aging and disease-a motivation for lower limb exoskeletons},
  author={Grimmer, Martin and Riener, Robert and Walsh, Conor James and Seyfarth, Andr{\'e}},
  journal={Journal of neuroengineering and rehabilitation},
  volume={16},
  number={1},
  pages={2},
  year={2019},
  publisher={Springer}
}

@article{freiberger2020mobility,
  title={Mobility in older community-dwelling persons: a narrative review},
  author={Freiberger, Ellen and Sieber, Cornel Christian and Kob, Robert},
  journal={Frontiers in physiology},
  volume={11},
  pages={881},
  year={2020},
  publisher={Frontiers Media SA}
}

@article{song2021deep,
  title={Deep reinforcement learning for modeling human locomotion control in neuromechanical simulation},
  author={Song, Seungmoon and Kidzi{\'n}ski, {\L}ukasz and Peng, Xue Bin and Ong, Carmichael and Hicks, Jennifer and Levine, Sergey and Atkeson, Christopher G and Delp, Scott L},
  journal={Journal of neuroengineering and rehabilitation},
  volume={18},
  number={1},
  pages={126},
  year={2021},
  publisher={Springer}
}

@article{schumacher2025emergence,
  title={Emergence of natural and robust bipedal walking by learning from biologically plausible objectives},
  author={Schumacher, Pierre and Geijtenbeek, Thomas and Caggiano, Vittorio and Kumar, Vikash and Schmitt, Syn and Martius, Georg and Haeufle, Daniel FB},
  journal={iScience},
  volume={28},
  number={4},
  year={2025},
  publisher={Elsevier}
}

@article{badie2025bioinspired,
  title={Bioinspired morphology and task curricula for learning locomotion in bipedal muscle-actuated systems},
  author={Badie, Nadine and Al-Hafez, Firas and Schumacher, Pierre and Haeufle, Daniel FB and Peters, Jan and Schmitt, Syn},
  journal={Communications Engineering},
  volume={4},
  number={1},
  pages={115},
  year={2025},
  publisher={Nature Publishing Group UK London}
}

@article{berg2024sar,
  title={Sar: Generalization of physiological agility and dexterity via synergistic action representation},
  author={Berg, Cameron and Caggiano, Vittorio and Kumar, Vikash},
  journal={Autonomous Robots},
  volume={48},
  number={8},
  pages={28},
  year={2024},
  publisher={Springer}
}

@article{he2024dynsyn,
  title={DynSyn: Dynamical synergistic representation for efficient learning and control in overactuated embodied systems},
  author={He, Kaibo and Zuo, Chenhui and Ma, Chengtian and Sui, Yanan},
  journal={arXiv preprint arXiv:2407.11472},
  year={2024}
}

@article{simos2025reinforcement,
  title={Reinforcement learning-based motion imitation for physiologically plausible musculoskeletal motor control},
  author={Simos, Merkourios and Chiappa, Alberto Silvio and Mathis, Alexander},
  journal={arXiv preprint arXiv:2503.14637},
  year={2025}
}

@inproceedings{park2025magnet,
  title={MAGNET: Muscle Activation Generation Networks for Diverse Human Movement},
  author={Park, Jungnam and Jung, Euikyun and Lee, Jehee and Won, Jungdam},
  booktitle={Proceedings of the Special Interest Group on Computer Graphics and Interactive Techniques Conference Conference Papers},
  pages={1--11},
  year={2025}
}

@article{davella2003combinations,
  title={Combinations of muscle synergies in the construction of a natural motor behavior},
  author={d'Avella, Andrea and Saltiel, Philippe and Bizzi, Emilio},
  journal={Nature neuroscience},
  volume={6},
  number={3},
  pages={300--308},
  year={2003},
  publisher={Nature Publishing Group US New York}
}

@article{delp2007opensim,
  title={OpenSim: open-source software to create and analyze dynamic simulations of movement},
  author={Delp, Scott L and Anderson, Frank C and Arnold, Allison S and Loan, Peter and Habib, Ayman and John, Chand T and Guendelman, Eran and Thelen, Darryl G},
  journal={IEEE transactions on biomedical engineering},
  volume={54},
  number={11},
  pages={1940--1950},
  year={2007},
  publisher={IEEE}
}

@article{dembia2020opensim,
  title={Opensim moco: Musculoskeletal optimal control},
  author={Dembia, Christopher L and Bianco, Nicholas A and Falisse, Antoine and Hicks, Jennifer L and Delp, Scott L},
  journal={PLOS Computational Biology},
  volume={16},
  number={12},
  pages={e1008493},
  year={2020},
  publisher={Public Library of Science San Francisco, CA USA}
}

@inproceedings{haarnoja2018soft,
  title={Soft actor-critic: Off-policy maximum entropy deep reinforcement learning with a stochastic actor},
  author={Haarnoja, Tuomas and Zhou, Aurick and Abbeel, Pieter and Levine, Sergey},
  booktitle={International conference on machine learning},
  pages={1861--1870},
  year={2018},
  organization={Pmlr}
}

@article{stable-baselines3,
  author  = {Antonin Raffin and Ashley Hill and Adam Gleave and Anssi Kanervisto and Maximilian Ernestus and Noah Dormann},
  title   = {Stable-Baselines3: Reliable Reinforcement Learning Implementations},
  journal = {Journal of Machine Learning Research},
  year    = {2021},
  volume  = {22},
  number  = {268},
  pages   = {1-8},
  url     = {http://jmlr.org/papers/v22/20-1364.html}
}

@article{scherpereel2023human,
  title={A human lower-limb biomechanics and wearable sensors dataset during cyclic and non-cyclic activities},
  author={Scherpereel, Keaton and Molinaro, Dean and Inan, Omer and Shepherd, Max and Young, Aaron},
  journal={Scientific Data},
  volume={10},
  number={1},
  pages={924},
  year={2023},
  publisher={Nature Publishing Group UK London}
}

@article{carmargo2021,
title = {A comprehensive, open-source dataset of lower limb biomechanics in multiple conditions of stairs, ramps, and level-ground ambulation and transitions},
journal = {Journal of Biomechanics},
volume={119},
pages = {110320},
year = {2021},
issn = {0021-9290},
doi = {https://doi.org/10.1016/j.jbiomech.2021.110320},
url = {https://www.sciencedirect.com/science/article/pii/S0021929021001007},
author = {Jonathan Camargo and Aditya Ramanathan and Will Flanagan and Aaron Young},
}

@article{cotton2025kintwin,
  title={KinTwin: Imitation Learning with Torque and Muscle Driven Biomechanical Models Enables Precise Replication of Able-Bodied and Impaired Movement from Markerless Motion Capture},
  author={Cotton, R James},
  journal={arXiv preprint arXiv:2505.13436},
  year={2025}
}

@article{weng2021natural,
  title={Natural walking with musculoskeletal models using deep reinforcement learning},
  author={Weng, Jiacheng and Hashemi, Ehsan and Arami, Arash},
  journal={IEEE Robotics and Automation Letters},
  volume={6},
  number={2},
  pages={4156--4162},
  year={2021},
  publisher={IEEE}
}

@article{defour2025impaired,
  title={Impaired motor control during post-stroke walking: a systematic review and meta-analysis of muscle synergies across different phases of recovery.},
  author={Defour, Arne and Dominici, Nadia and Swinnen, Eva and Cambier, Dirk and Van Cleemput, Gitte and Van Bladel, Anke},
  journal={Gait \& posture},
  pages={109978},
  year={2025},
  publisher={Elsevier}
}

@article{clark2010merging,
  title={Merging of healthy motor modules predicts reduced locomotor performance and muscle coordination complexity post-stroke},
  author={Clark, David J and Ting, Lena H and Zajac, Felix E and Neptune, Richard R and Kautz, Steven A},
  journal={Journal of neurophysiology},
  volume={103},
  number={2},
  pages={844--857},
  year={2010},
  publisher={American Physiological Society Bethesda, MD}
}

@article{van2020lower,
  title={Lower limb muscle synergies during walking after stroke: a systematic review},
  author={Van Criekinge, Tamaya and Vermeulen, Jordi and Wagemans, Keanu and Schr{\"o}der, Jonas and Embrechts, Elissa and Truijen, Steven and Hallemans, Ann and Saeys, Wim},
  journal={Disability and rehabilitation},
  volume={42},
  number={20},
  pages={2836--2845},
  year={2020},
  publisher={Taylor \& Francis}
}

@article{berniker2009simplified,
  title={Simplified and effective motor control based on muscle synergies to exploit musculoskeletal dynamics},
  author={Berniker, Max and Jarc, Anthony and Bizzi, Emilio and Tresch, Matthew C},
  journal={Proceedings of the National Academy of Sciences},
  volume={106},
  number={18},
  pages={7601--7606},
  year={2009},
  publisher={National Academy of Sciences}
}

@article{bizzi2013neural,
  title={The neural origin of muscle synergies},
  author={Bizzi, Emilio and Cheung, Vincent CK},
  journal={Frontiers in computational neuroscience},
  volume={7},
  pages={51},
  year={2013},
  publisher={Frontiers Media SA}
}

@article{bernstein1967coordination,
  title={The coordination and regulation of movements},
  author={Bernstein, Nicholas},
  journal={(No Title)},
  year={1967}
}

@article{d2005shared,
  title={Shared and specific muscle synergies in natural motor behaviors},
  author={d'Avella, Andrea and Bizzi, Emilio},
  journal={Proceedings of the national academy of sciences},
  volume={102},
  number={8},
  pages={3076--3081},
  year={2005},
  publisher={National Academy of Sciences}
}

@article{d2006control,
  title={Control of fast-reaching movements by muscle synergy combinations},
  author={d'Avella, Andrea and Portone, Alessandro and Fernandez, Laure and Lacquaniti, Francesco},
  journal={Journal of Neuroscience},
  volume={26},
  number={30},
  pages={7791--7810},
  year={2006},
  publisher={Society for Neuroscience}
}

\end{document}